\documentclass[journal]{IEEEtran}

\usepackage{graphicx}
\usepackage{subfigure}
\usepackage{cite}
\usepackage[T1]{fontenc}
\usepackage{float}
\usepackage{stfloats}
\usepackage[colorlinks,linkcolor=black,citecolor=black]{hyperref}

\ifCLASSINFOpdf
\else
\fi

\begin{document}
\title{DDU-Net: Dual-Decoder-U-Net for Road Extraction Using High-Resolution Remote Sensing Images}

\author{Ying~Wang,
        Yuexing~Peng,~\IEEEmembership{Member,~IEEE,}
        Xinran~Liu,
        Wei~Li,~\IEEEmembership{Senior~Member,~IEEE,}
        George~C.~Alexandropoulos,~\IEEEmembership{Senior~Member,~IEEE,}\\
        Junchuan~Yu,~Daqing~Ge,
        and Wei~Xiang,~\IEEEmembership{Senior~Member,~IEEE}
\thanks{Y. Wang and Y. Peng are with the School of Information and Communication Engineering, Beijing University of Posts and Telecommunications, 100876, Beijing, P.R.China (email: \{wangying\_0325, yxpeng, lxr2019\}@bupt.edu.cn).}
\thanks{W. Li is with the School of Information and Electronics, Beijing Institute of Technology, 100081, Beijing, P.R China (email: liwei089@ieee.org).}
\thanks{G. C. Alexandropoulos is with the Department of Informatics and Telecommunications, National and Kapodistrian University of Athens, Panepistimiopolis Ilissia, 15784 Athens, Greece (email: alexandg@di.uoa.gr).}
\thanks{J. Yu and D. Ge are with China Aero Geophysical Survey and Remote Sensing Center for Natural Resources, Beijing, 10083, China (email: \{yujunchuan, gedaqing\}@mail.cgs.gov.cn).}
\thanks{W. Xiang is with the School of Engineering and Mathematical Sciences, La Trobe University, Melbourne, Australia (email: w.xiang@latrobe.edu.au).}
\thanks{This work is supported by National Key Research and Development Program of China under grant 2021YFC3000400. (Corresponding author: Yuexing Peng.)}}


\maketitle

\begin{abstract}
Extracting roads from high-resolution remote sensing images (HRSIs) is vital in a wide variety of applications, such as autonomous driving, path planning, and road navigation. Due to the long and thin shape as well as the shades induced by vegetation and buildings, small-sized roads are more difficult to discern. In order to improve the reliability and accuracy of small-sized road extraction when roads of multiple sizes coexist in an HRSI, an enhanced deep neural network model termed Dual-Decoder-U-Net (DDU-Net) is proposed in this paper. Motivated by the U-Net model, a small decoder is added to form a dual-decoder structure for more detailed features. In addition, we introduce the dilated convolution attention module (DCAM) between the encoder and decoders to increase the receptive field as well as to distill multi-scale features through cascading dilated convolution and global average pooling. The convolutional block attention module (CBAM) is also embedded in the parallel dilated convolution and pooling branches to capture more attention-aware features. Extensive experiments are conducted on the Massachusetts Roads dataset with experimental results showing that the proposed model outperforms the state-of-the-art DenseUNet, DeepLabv3+ and D-LinkNet by 6.5\%, 3.3\%, and 2.1\% in the mean Intersection over Union (mIoU), and by 4\%, 4.8\%, and 3.1\% in the F1 score, respectively. Both ablation and heatmap analyses are presented to validate the effectiveness of the proposed model.
\end{abstract}


\section{Introduction}

\IEEEPARstart{W}{ith}  rapid development of remote sensing technology, high-resolution remote sensing images (HRSIs) are of great significance in wide-ranging application areas, such as land use and urban planning\cite{transferring}. Road extraction from HRSIs has always attracted much research interest from both academics and industries in areas such as autonomous driving, path planning, road navigation, intelligent transport, and emergency support\cite{centerline,automatic,graph-based}.

Roads in HRSIs exhibit the following typical features\cite{road_features}: (1) Geometric features: roads are of long and thin shape; (2) Radiation feature: a road has two obvious edges, and its inner gray level often differs greatly from adjacent areas; (3) Topological feature: roads are interconnected to form a network; (4) Context features: roads, roadside buildings, and street trees constitute the local context features, while the surrounding urban or rural region comprise the global context feature. Considering these features, many road extraction models have been proposed, which can be roughly categorized into traditional models and deep convolutional neural network (DCN)-based models. The traditional ones mainly exploit the geometric features, such as edge detection\cite{edge_detection}, snake model\cite{snake1,snake2,snake3}, and dynamic programming\cite{dynamic_programming}. The DCN-based models include semantic segmentation-type models, like FCN\cite{FCN}, U-Net\cite{U-Net}, LinkNet\cite{LinkNet}, PSPNet\cite{pspnet}, and DeepLab\cite{deeplab}, as well as their variants, such as D-LinkNet\cite{dlinknet}, DeepLabv3+\cite{deeplabv3+}, and HsgNet\cite{hsgnet}, which have become the cutting-edge models with the state-of-the-art performance.

\begin{figure}[tbp]
	\centering
	\label{fig1}
	\subfigure[]{
		\begin{minipage}[t]{0.35\linewidth}
			\centering
			\includegraphics[width=\linewidth]{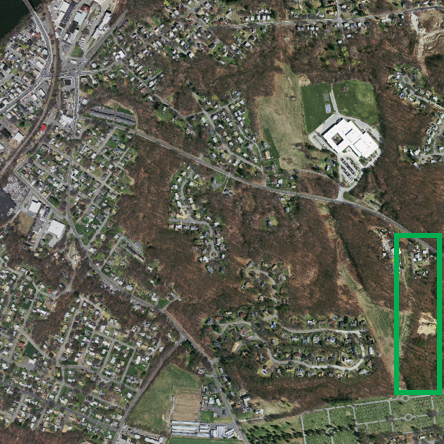}
			\\[5pt]
			\includegraphics[width=\linewidth]{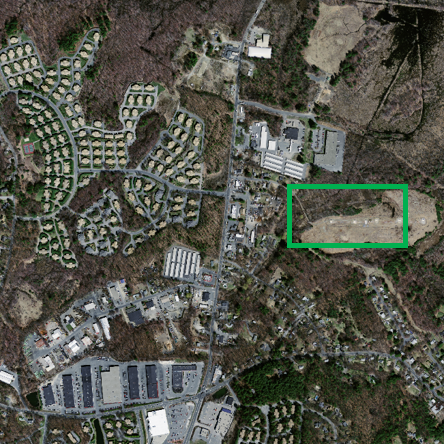}
			\vspace{-7pt}
		\end{minipage}	
	}
    \hspace{-0.025\linewidth}
	\subfigure[]{
		\begin{minipage}[t]{0.35\linewidth}
			\centering
			\includegraphics[width=\linewidth]{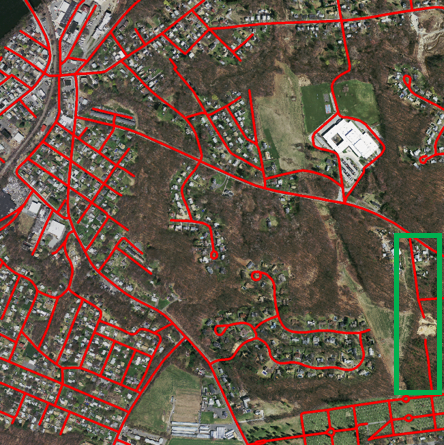}
			\\[5pt]
			\includegraphics[width=\linewidth]{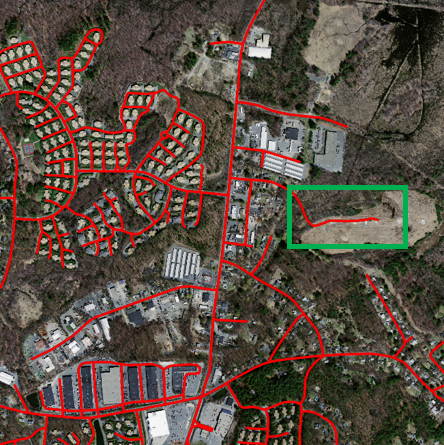}
			\vspace{-7pt}
		\end{minipage}
	}
	\caption{Typical challenges in road segmentation. (a) samples of HRSIs from the Massachusetts Roads dataset, (b) road segmentation mask (ground truth). The green boxes highlight small-sized roads vulnerable to interferences like shadows of vegetation, in the first case, and similarity to the background, in color in the second case.}
	\vspace{-1em}
\end{figure}
There still exist some big challenges for reliable road segmentation from HRSIs, including: (1) small-sized roads usually only take up a very small proportion in a road image resulting in a severe imbalance between foreground and background; (2) roads hold geometric similarity to rivers and gullies so that it is prone to misclassification; (3) the low spatial resolution of HRSIs induces mixed pixels of roads, such as vehicles on the road and the shadow of roadside vegetation or buildings; (4) small-sized roads are more vulnerable to interference, resulting in missing detection and discontinuity. As illustrated in Fig.~\ref{fig1}, one of the most common problems in road extraction is related to tree shadows, which lead to road discontinuity (an example can be seen in the first case). Another problem is related to the similarity between a road and its surrounding ground, leading to missing detection and misclassification (as shown in the second case).

In view of the aforementioned challenges, an enhanced road extraction model termed Dual-Decoder-U-Net (DDU-Net) is developed in this paper with the objective of enhancing the performance of small-sized road extraction. The essential ideas behind DDU-Net are: (1) extracting and fusing multi-scale features by the skip connection between the encoder and decoder for tackling the problem of road extraction with different sizes; (2) expanding the receptive field without loss of detailed information by dilated convolution; (3) distilling context information by the channel spatial attention mechanism; and (4) restoring more details by the dual decoder structure via adding a small decoder after three rounds of down-sampling to utilize more detailed features.

The main contributions can be summarized as follows.

1)	A dual-decoder U-Net like semantic segmentation model termed DDU-Net is proposed to greatly enhance the road extraction performance in complex environments. In the proposed DDU-Net model, a dual-decoder structure is designed by embedding a small decoder before the decoder for retaining more detailed information, which can greatly improve the performance of road extraction, especially for small-sized roads due to its susceptibility to inference of roadside vegetation and buildings;

2) A dilated convolution attention module (DCAM) is introduced after the encoder to expand the receptive field. In the DCAM, multi-scale features are obtained by stacking dilated convolutions and introducing global average pooling, and the convolutional block attention module (CBAM)\cite{cbam} is employed to capture attention-aware features through the channel spatial attention mechanism; and

3)	Performance evaluation experiments on the open Massachusetts Roads dataset show that the proposed DDU-Net model can improve the performance of road extraction, especially for small-sized roads, and outperforms its state-of-the-art DeepLabv3+, DenseUNet and D-LinkNet counterparts. Through ablation experiments and a heatmap analysis, the introduction of the DCAM and the small decoder before the decoder is proven to be able to markedly enhance the model's capability of learning more details and then improve road extraction in complex environments.

\section{Related Work}
Many works have been carried out on road extraction from remote sensing images. Early works mainly utilize radiation features, geometric features, and topological features of roads. For example, a wavelet function is employed in \cite{edge_detection} for road edge detection, the snake algorithm is used to extract road contour in \cite{snake1,snake2,snake3}, and the Hough line detection method is employed by \cite{hough} to catch straight line segments and then connect them to form road curves. In \cite{road_features}, a parallel double-edge method is designed to determine the road area by virtual of road geometric features.

With the explosive development of deep learning, the road extraction task is being performed as a problem of pixel-level classification by use of convolutional neural networks (CNNs)\cite{line_integral_convolution,roadtracer,pspnet,ResUnet,stacked_unets,dlinknet,hsgnet,fcn_ensemble}. In \cite{line_integral_convolution}, the CNN is used to predict the probability of pixels belonging to roads, and then the line integral convolution based smoothing algorithm is designed for small gap connection. In \cite{roadtracer}, an iterative search method is proposed on top of the CNN decision function to avoid complicated post-processing. Later, semantic segmentation models are applied to road extraction, including FCN\cite{FCN} featuring convolution instead of the fully-connected layer, U-Net\cite{U-Net} with a simple encoder-decoder structure and skip connection, LinkNet\cite{LinkNet} with high-efficiency computing and storage, PSPNet\cite{pspnet} with pyramid pooling for multi-scale features extraction, and DeepLabv3+\cite{deeplabv3+} combining spatial pyramid pooling and encoder-decoder structure. Also, some improved models are developed to fully distill road characteristics. For example, ResUnet\cite{ResUnet} replaces the common convolution layer by a residual unit in the encoder of U-Net; Stacked U-Nets\cite{stacked_unets} stack two U-Nets to yield road structure features and segmentation mask, respectively; D-LinkNet\cite{dlinknet} improves the original LinkNet by adding a cascaded dilated convolution module to expand the receptive field and to extract context information; HsgNet\cite{hsgnet} substitutes the spatial attention module for the dilated convolution module on the basis of D-LinkNet to make full use of global spatial information; based on FCN, an ensemble learning model is proposed in \cite{fcn_ensemble} to integrate the road extraction results using several weighted loss functions; DenseUNet\cite{denseunet} consists of dense connection units and skip connections to deal with the problem of tree and shadow occlusion.

Although many multi-resolution techniques have been developed for multi-scale feature extraction such as the encoder and decoder in U-Net, spatial pyramid pooling in PSPNet, and dilated convolution in D-LinkNet, these techniques focus mainly on context information extraction from the feature map after multiple down sampling operations, while small-sized roads still have high missing rates and small-sized road extraction is still a research area worth exploring.

In this paper, we propose an enhanced road extraction model dubbed DDU-Net to generate improved road semantic segmentation predictions especially for small-sized roads. The model combines dilated convolution and an attention mechanism, aiming at enhancing global context semantic feature extraction. Moreover, by adopting the dual-decoder structure, DDU-Net can retain more low-level features and thus present more details for small-sized road detection.

\section{Proposed Method}
\begin{figure*}[htb]
	\centering
	\includegraphics[width=0.78\textwidth]{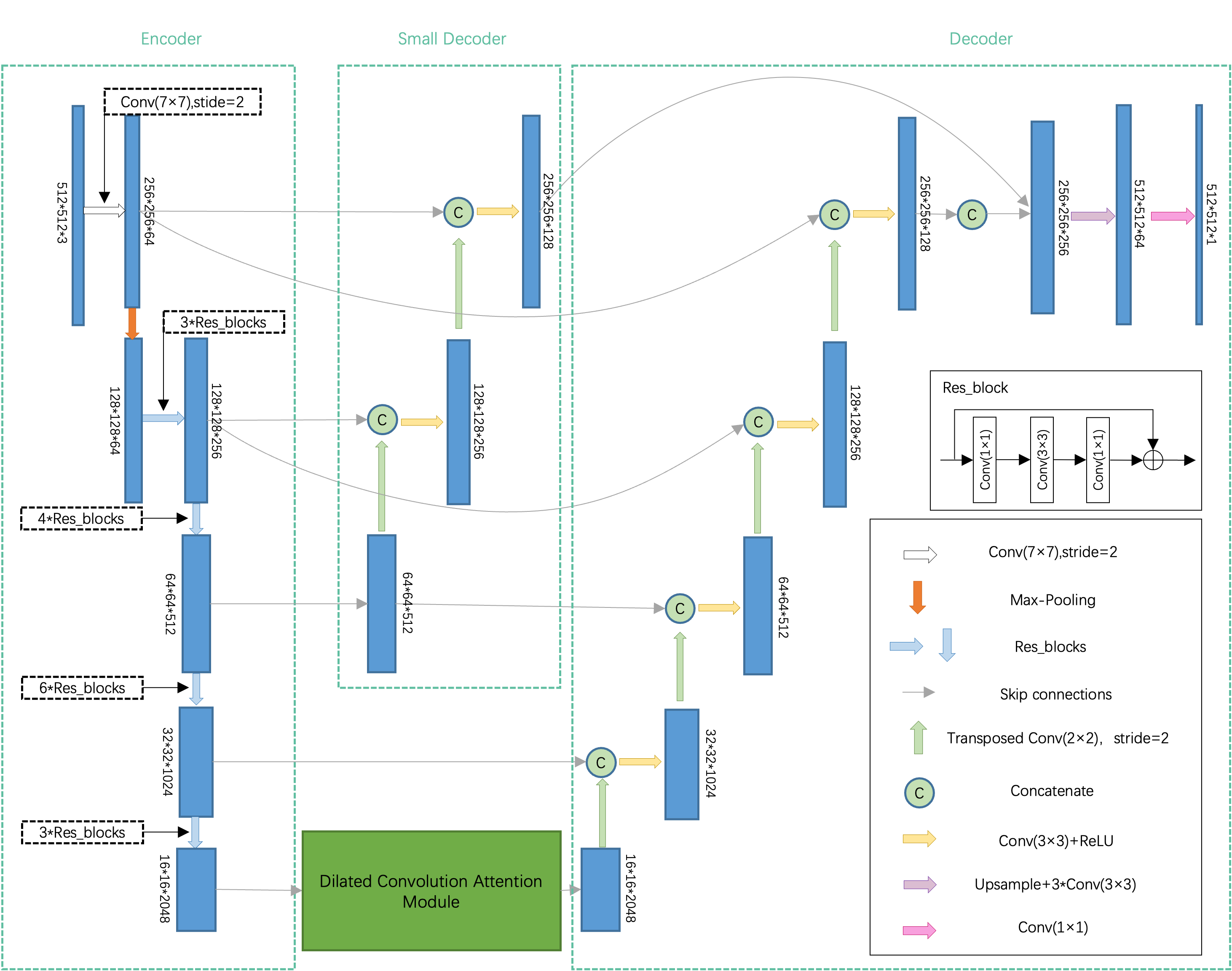}
	\caption{Network architecture of the DDU-Net model, which consists of an encoder, a dilated convolution attention module (DCAM), and a dual-decoder.}
	\label{fig:figure2}
\end{figure*}

As shown in Fig. \ref{fig:figure2}, motivated by U-Net, the proposed DDU-Net model consists of an encoder, the dilated convolution attention module (DCAM), and a dual-decoder, where the encoder extracts feature maps and the decoder maps road features into spatial areas for segmentation.

\begin{figure}[tbp]
	\centering
	\includegraphics[width=\linewidth]{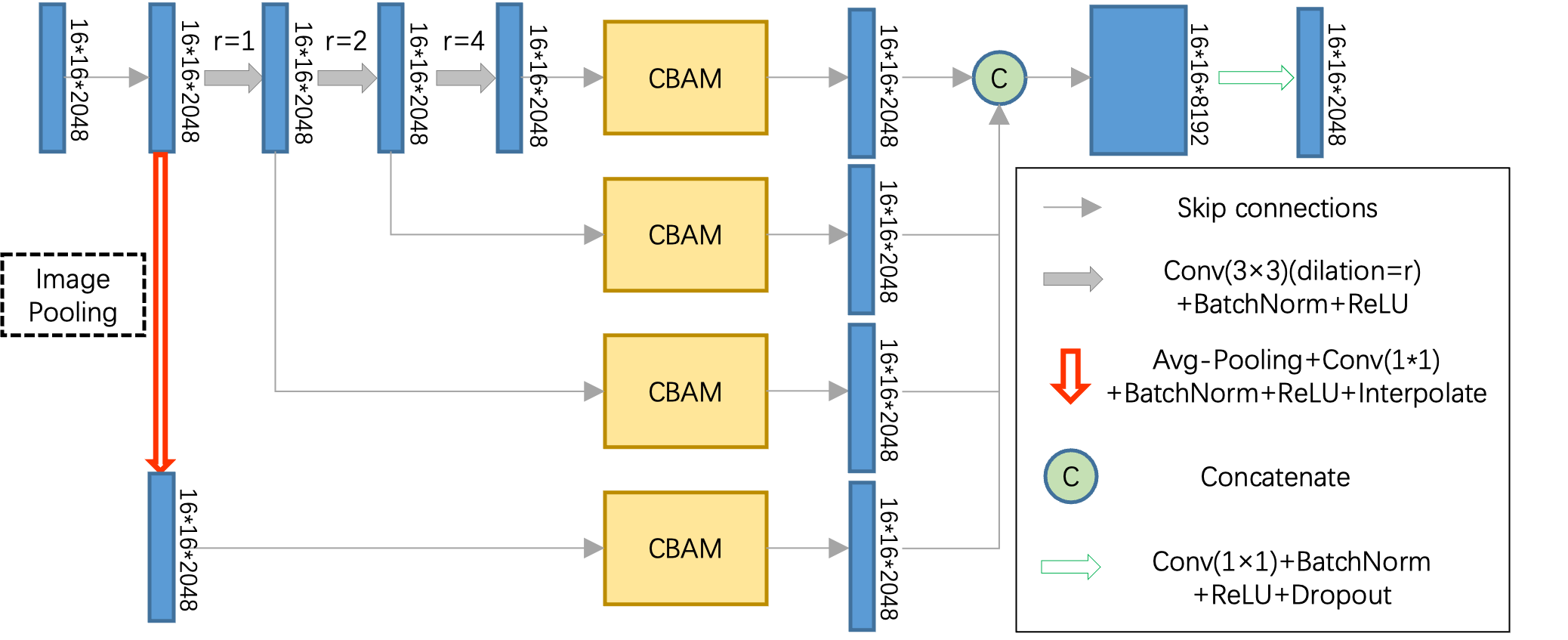}
	\caption{Schematics of the DCAM.}
	\label{fig:figure3}
	\vspace{-1em}
\end{figure}
\begin{figure}[tbp]
	\centering
	\includegraphics[width=\linewidth]{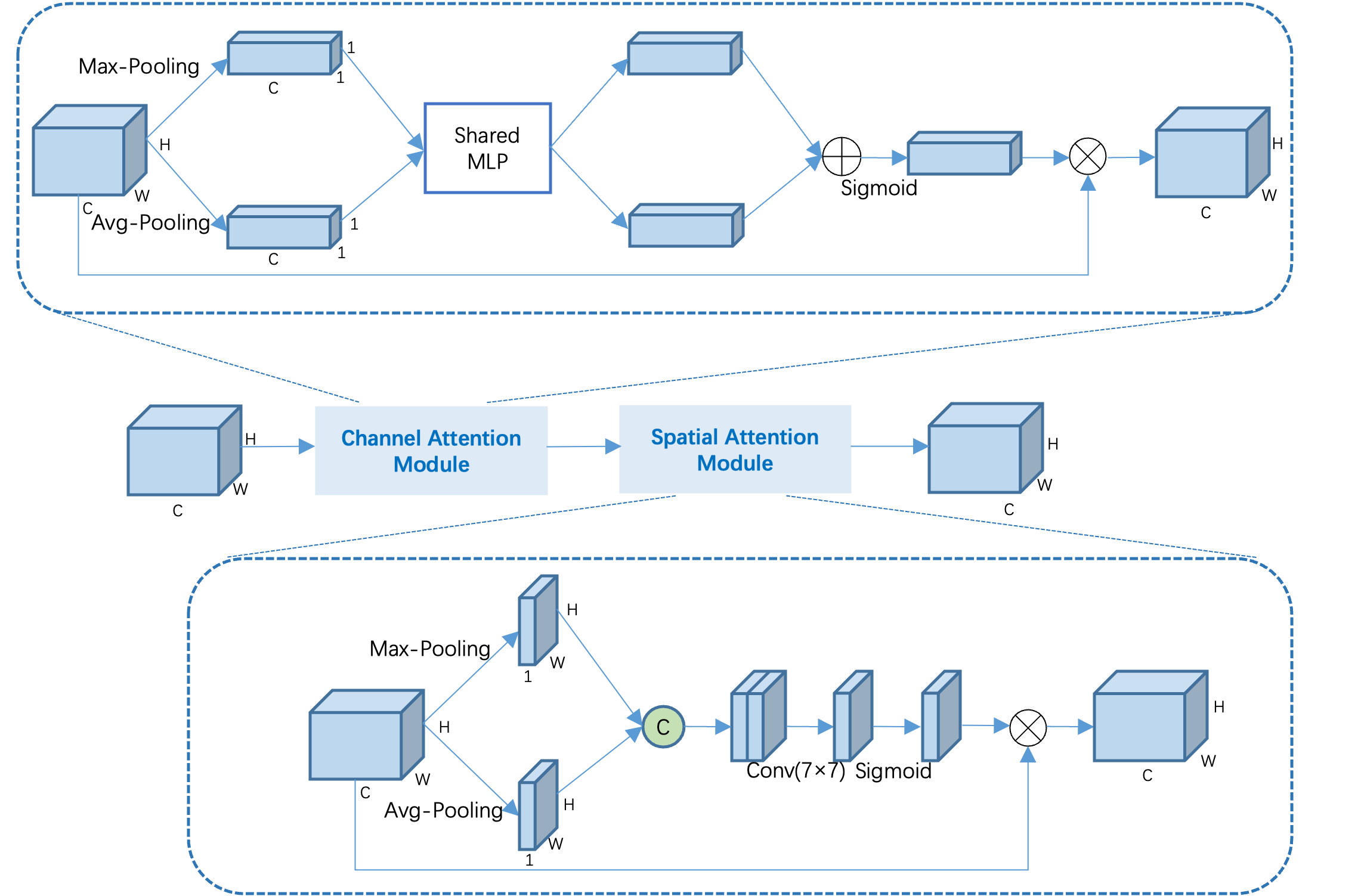}
	\caption{Schematics of the CBAM \cite{cbam}.}
	\label{fig:figure_cbam}
	\vspace{-1em}
\end{figure}

\subsection{Encoder}
DDU-Net uses ResNet-50 pretrained on the ImageNet dataset as its encoder, which is then fine-tuned on the Massachusetts Roads Dataset, so as to accelerate the convergence and achieve better performance via transfer learning. The multi-scale feature maps are fused by the skip connection to enhance the proposed model's generalization ability. We focus more on the specific feature extraction layer, which plays a decisive role in the decoder stage, and design the dual decoder architecture. Meanwhile, considering the long span and connectivity of roads, we introduce the dilated convolution attention module (DCAM).

\subsection{Dilated Convolution Attention Module (DCAM)}


The DCAM is introduced to expand the receptive field. As depicted in Fig. \ref{fig:figure3}, the DCAM has four parallel branches with four reception sizes. The first three branches are cascaded to fuse feature maps of different receptive field sizes, while the last branch adds image-level features via global average pooling. As shown in Fig. \ref{fig:figure_cbam}, the convolutional block attention module (CBAM) \cite{cbam} is embedded in each parallel branch to obtain attention-aware features, which are concatenated and reduced by 1 $\times$ 1 convolution to form features of the same dimension. The channel characteristic weight matrix $M_{C}$ and spatial characteristic weight matrix $M_{S}$ are defined as follows
\begin{equation}
	M_{C}(F)=\sigma(\textrm{MLP}( \textrm{AvgPool}(F))+\rm{MLP}(\rm{MaxPool}(F))),
\end{equation}
\begin{equation}
	M_{S}(F)=\sigma(f^{7\times7}([\rm{AvgPool}(F); \rm{MaxPool}(F)])),
\end{equation}
where $\sigma$ denotes the sigmoid activation function, $\rm{MLP}$ is a neural network with two layers, and $f^{7\times7}$ represents a convolution operation with the filter size of $7\times7$.

Dilated convolution can expand the receptive fields of the feature points without reducing the resolution of the feature maps. It requires neither additional parameters nor extra computation, and thus is widely used. The sizes of the receptive fields of the feature maps can be modified by adjusting the dilated rate. However, the gridding effect exists in dilated convolution, leading to a loss of local information and a degradation in small-sized object detection. Cascading dilated convolutions of differing rates is conducive to obtaining information from a wider range of pixels and avoding the gridding effect. In this work, the size is reduced from the original input images of 512 $\times$ 512 to 16 $\times$ 16 in five rounds of down-sampling in ResNet-50. Therefore, we cascade convolutions with dilated rates of 1, 2 and 4, and the receptive fields of parallel branches are 3, 7 and 15, respectively. In this way, the whole input feature map can be roughly covered. On this basis, an average pooling branch is added to obtain the global information of HRSIs. The CBAM performs both channel attention mapping and spatial attention mapping, where the weight coefficients are obtained by both the average-pooling and max-pooling. The channel spatial dual-attention mechanism helps the network autonomously learn feature weights and obtain attention-aware features of different receptive fields.

\subsection{Dual Decoder}
The decoder is designed to accommodate the dual-decoder structure in order to capture both large-scale and small-scale features for the enhancement of small-sized road extraction.

It is well known that the backbone network of the semantic segmentation model usually consists of multiple convolutional layers, which gradually abstract multi-scale features. High-level features indicate large receptive fields, strong representation ability of semantic information, low-resolution feature maps, and few details of spatial geometric features. By contrast, low-level features imply small receptive fields, high-resolution feature maps, weak representation of semantic information but strong representation of detail information. The semantic information of high-level features is conducive to accurate object detection, while the geometric information of low-level features helps restore more details.

In the encoder, ResNet-50 features a series of convolutional layers and adopts five rounds of down-sampling, which reduces the resolution and increases the receptive field of the feature map gradually. In the decoder, the process is reversed, and the representations learned are upsampled to the size of the input image finally through transposed convolutions. It is easy to miss small-sized objects if the decoder only uses the highest feature map of the encoder. Therefore, the U-Net model adds skip connections to enable the sharing of low-level information between the encoder and decoder so as to preserve low-level features and achieve multi-scale feature fusion. The U-Net model is designed originally for medical image segmentation, where cells have a fixed structure and similar sizes. Through the use of up-sampling, down-sampling and skip connection, the model can extract and fuse multi-scale features, so as to achieve good segmentation results. However, for the task of road extraction, the U-Net model behaves unsatisfactorily with a high missing rate and poor road continuity, especially for small-sized roads~\cite{improve-unet}. Since the sizes and shapes of roads vary greatly and the backgrounds are diversed, road segmentation is much different from and more difficult than cell segmentation. Therefore, we develop a dual-decoder structure consisting of a small decoder and a conventional full-size decoder. The small decoder is appended to the encoder, whose output in three rounds of down-sampling is up-sampled to the dimension of 256 $\times$ 256. The output features from both decoders are fused to generate feature maps of dimension 256 $\times$ 256 $\times$ 256.

By utilizing the first four levels of the feature map and also the output of the DCAM, the large decoder can distill multi-scale context information from the highest-level feature map, and then integrates global context semantic information efficiently. Meanwhile, the small decoder pays more attention to the restoration of edge details from low-level feature maps, and refines the outputs of the large decoder. This dual-decoder structure is able to effectively strength low-level features and enhance small-sized road extraction, which will be validated by experiments and a heatmap analysis in the ensuing section.

\section{Experimental Results and Discussions}

\subsection{Dataset}
We evaluate our model on the open Massachusetts Roads dataset\cite{Massachusetts}, which covers more than 2600 square kilometers of cities, suburbs and villages, including roads, and buildings, vegetation, vehicles, and so on. This open dataset consists of 1108 training images, 14 validation images, and 49 test images. The resolution of each image is 1500 $\times$ 1500, and the road resolution is 1 meter/pixel. The dataset is formulated as a binary segmentation problem, in which the pixel
value of 0 is assigned to pixels belonging to the background
class, and the pixel value of 1 is assigned to pixels belonging to
the road class.

For fair comparison, we follow the dataset division settings in \cite{Massachusetts}, where images in the training and validation sets are cropped with a step size of 484 pixels, and images in the test set are not cropped. Then the dataset contains 9972 training and 126 validation samples of size 512 $\times$ 512 pixels, and 49 test images of original size 1500 $\times$ 1500 pixels.

\subsection{Training Details}
The focal loss\cite{focal_loss} is employed as the loss function, which is defined as
\begin{equation}
L_{\rm{FL}}(p_t)=-(1-p_t)^\gamma \textrm{log} (p_t),
\end{equation}
where $p_t$ denotes the probability that the sample belongs to the true class, $\gamma=2$ to pay more attention to the samples that are difficult to classify. The experiment settings of the proposed model are listed in Table~\ref{table2}. All models are trained and tested on NVIDIA Tesla V100 SXM2 16GB.

\begin{table}[htb]
	\renewcommand\arraystretch{0.95}
	\centering
	\caption{Experiment settings}
	\label{table2}
	\resizebox{0.8\columnwidth}{!}{
	\begin{tabular}{l|l}
		\hline
		Item        & Parameter     \\
        \hline
		Optimizer   & Adam    \\
		Batch size  & 4   \\
		Initial learning rate & 0.001     \\
		Learning rate attenuation strategy & poly \\
        Attenuation coefficient & 0.9 \\
        Weight attenuation & $5\times e^{-4}$  \\
        Epoch & 50 \\
        \hline
	\end{tabular}
	}
\end{table}

\subsection{Performance Metrics}
We use the mean Intersection over Union (mIoU)\cite{miou} as the performance metric, which denotes the pixel-wise intersection over union score (IoU)\cite{miou} of each class. They are defined as
\begin{equation}
\textrm{mIoU}=\frac{1}{K+1} \sum_{i=0}^K \textrm{IoU}_i,
\end{equation}

\begin{equation}
\textrm{IoU}_i=\frac{\textrm{TP}_i}{\textrm{TP}_i+\textrm{FP}_i+\textrm{FN}_i},
\end{equation}
where TP, FP and FN represent the true positive, false positive, and false negative, respectively. When there are $K$ classes, mIoU is the average of IoU for all classes. For the current task, there are only two classes, i.e., road (class 1) and background (class 0). Since only the road class is taken into consideration and the background class is ignored, mIoU is identical to the IoU of the road class.

In addition, more comprehensive evaluation metrics, i.e., accuracy, precision, recall and F1 score, are also evaluated, which are defined as

\begin{equation}
\textrm{Accuracy}=\frac{\textrm{TP}+\textrm{FN}}{\textrm{TP}+\textrm{TN}+\textrm{FP}+\textrm{FN}},
\end{equation}

\begin{equation}
\textrm{Precision}=\frac{\textrm{TP}}{\textrm{TP}+\textrm{FP}},
\end{equation}

\begin{equation}
	\textrm{Recall}=\frac{\textrm{TP}}{\textrm{TP}+\textrm{FN}},
\end{equation}

\begin{equation}
	\textrm{F1\_score}=\frac{2 \times \textrm{Precision} \times \textrm{Recall}}{\textrm{Precision+Recall}}.
\end{equation}

\subsection{Results and Discussions}
Table~\ref{table1} lists the quantitative experimental results of four evaluated models on the Massachusetts Roads dataset. Since there exist severe class unbalance in the road extraction task, the F1 score and mIoU are more appropriate. As can be observed from Table \ref{table1}, the proposed DDU-Net model outperforms DeepLabv3+, DenseUNet and D-LinkNet by 4.8\%, 4.0\% and 3.1\% in F1 score, by 3.3\%, 6.5\% and 2.1\% in mIoU, respectively. Although its precision metric is worse than that of D-LinkNet, DDU-Net performs much better in terms of accuracy, recall, F1 score and mIoU. It can be concluded that the overall performance of DDU-Net is better than D-LinkNet and the other two mainstream models.

\begin{figure*}[htbp]
	\centering
	\subfigure[]{
		\begin{minipage}[t]{0.15\linewidth}
			\centering
			\includegraphics[width=\linewidth]{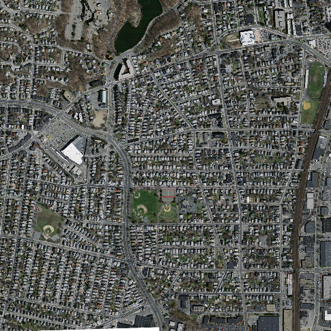}
			\\[5pt]
			\includegraphics[width=\linewidth]{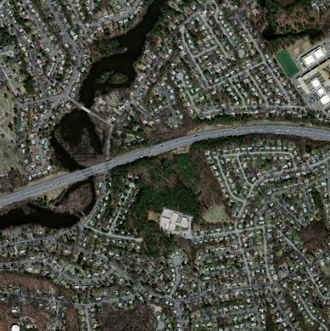}
			\\[5pt]
			\includegraphics[width=\linewidth]{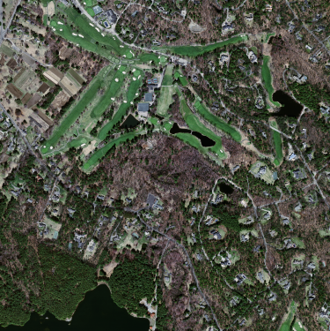}
			\\[5pt]
			\includegraphics[width=\linewidth]{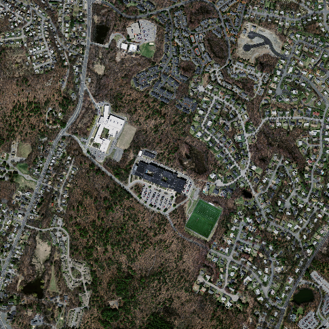}
			\\[5pt]
			\includegraphics[width=\linewidth]{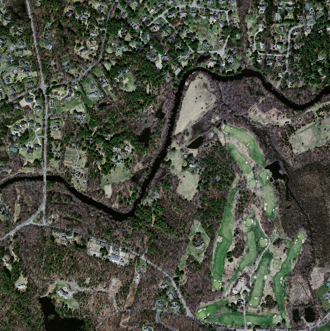}
			\\[5pt]
			\includegraphics[width=\linewidth]{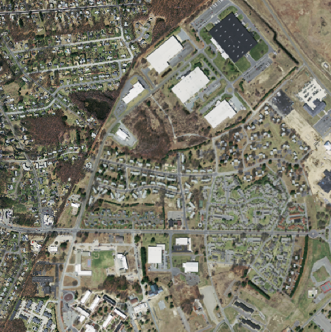}
			\vspace{-7pt}
		\end{minipage}	
	}
    \hspace{-0.01\linewidth}
	\subfigure[]{
		\begin{minipage}[t]{0.15\linewidth}
			\centering
			\includegraphics[width=\linewidth]{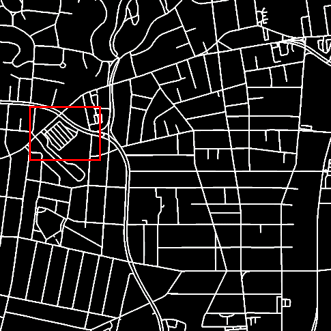}
			\\[5pt]
			\includegraphics[width=\linewidth]{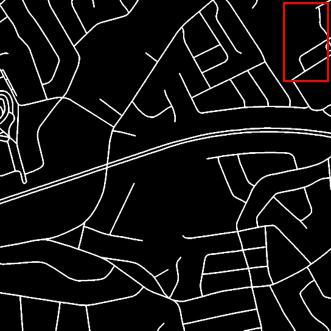}
			\\[5pt]
			\includegraphics[width=\linewidth]{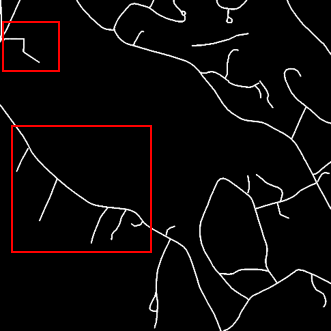}
			\\[5pt]
			\includegraphics[width=\linewidth]{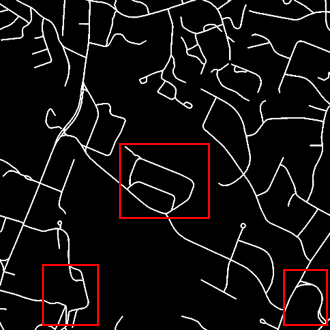}
			\\[5pt]
			\includegraphics[width=\linewidth]{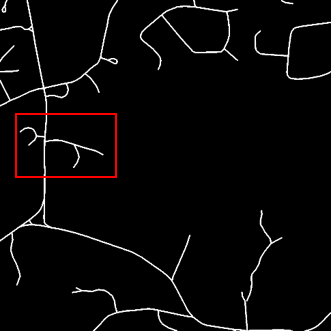}
			\\[5pt]
			\includegraphics[width=\linewidth]{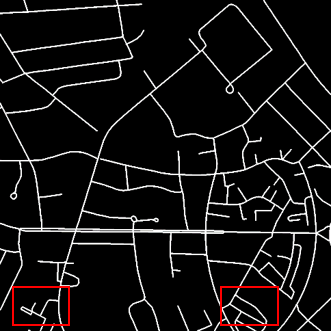}
			\vspace{-7pt}
		\end{minipage}
	}
    \hspace{-0.01\linewidth}
	\subfigure[]{
		\begin{minipage}[t]{0.15\linewidth}
			\centering
			\includegraphics[width=\linewidth]{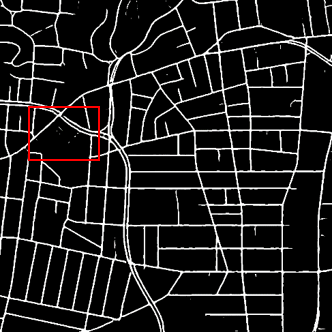}
			\\[5pt]
			\includegraphics[width=\linewidth]{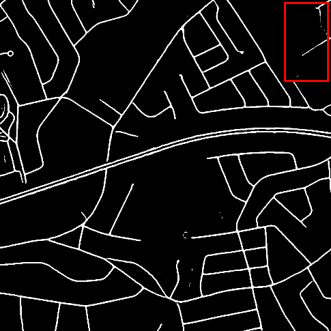}
			\\[5pt]
			\includegraphics[width=\linewidth]{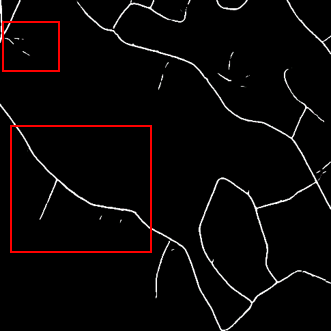}
			\\[5pt]
			\includegraphics[width=\linewidth]{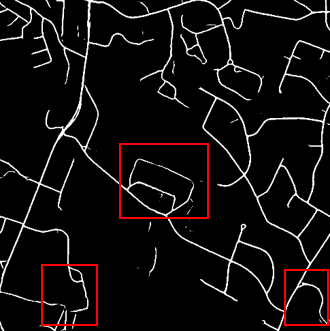}
			\\[5pt]
			\includegraphics[width=\linewidth]{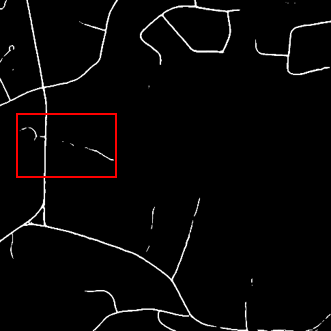}
			\\[5pt]
			\includegraphics[width=\linewidth]{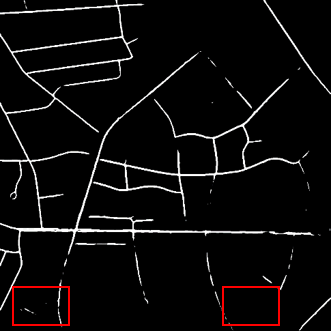}
			\vspace{-7pt}
		\end{minipage}
	}
    \hspace{-0.01\linewidth}
	\subfigure[]{
		\begin{minipage}[t]{0.15\linewidth}
			\centering
			\includegraphics[width=\linewidth]{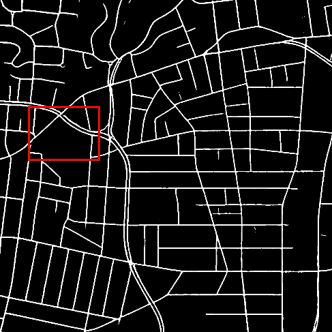}
			\\[5pt]
			\includegraphics[width=\linewidth]{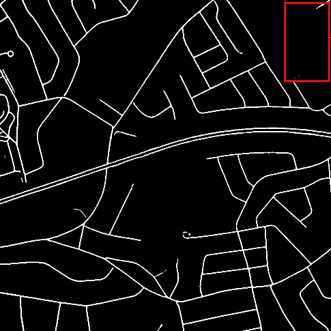}
			\\[5pt]
			\includegraphics[width=\linewidth]{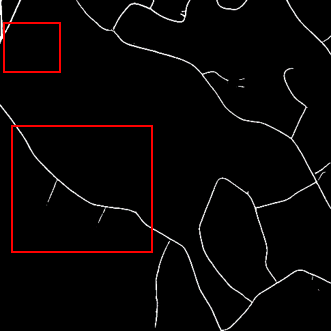}
			\\[5pt]
			\includegraphics[width=\linewidth]{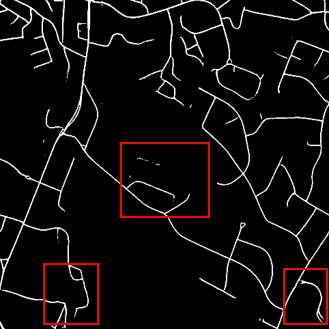}
			\\[5pt]
			\includegraphics[width=\linewidth]{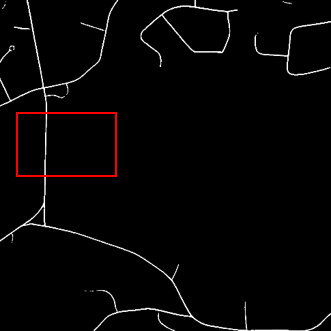}
			\\[5pt]
			\includegraphics[width=\linewidth]{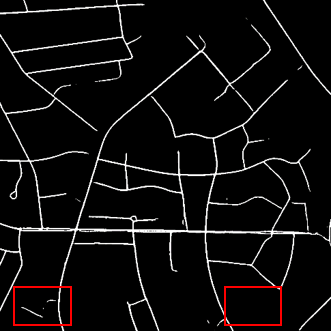}
			\vspace{-7pt}
		\end{minipage}
	}
    \hspace{-0.01\linewidth}
	\subfigure[]{
		\begin{minipage}[t]{0.15\linewidth}
			\centering
			\includegraphics[width=\linewidth]{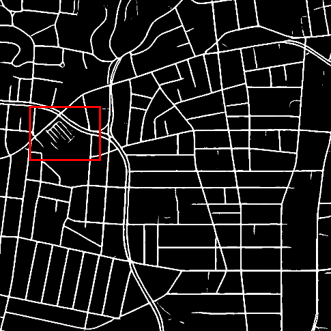}
			\\[5pt]
			\includegraphics[width=\linewidth]{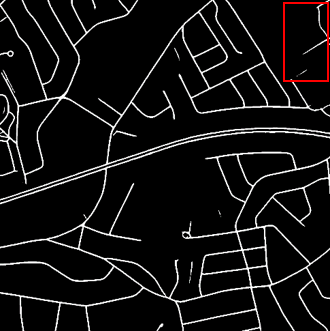}
			\\[5pt]
			\includegraphics[width=\linewidth]{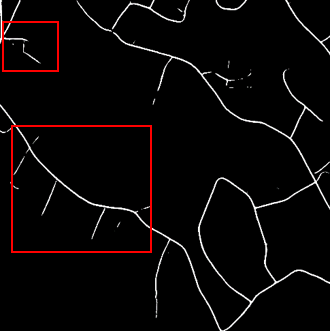}
			\\[5pt]
			\includegraphics[width=\linewidth]{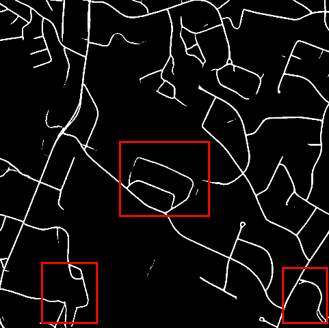}
			\\[5pt]
			\includegraphics[width=\linewidth]{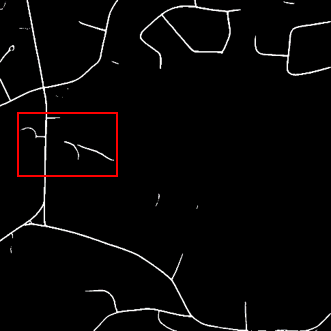}
			\\[5pt]
			\includegraphics[width=\linewidth]{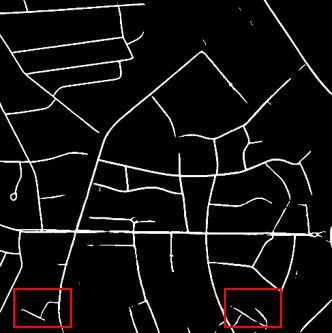}
			\vspace{-7pt}
		\end{minipage}
	}	
	\caption{Visual experimental results on the Massachusetts test set. (a) input image, (b) ground truth, (c)  DeepLabv3+, (d) D-LinkNet, and (e) ours. The red rectangles highlight the small-sized roads that our model has a significant improvement over the reference ones.}
	\label{fig:figure4}
	\vspace{-1em}
\end{figure*}

\begin{table}[t]
	\renewcommand\arraystretch{1.2}
	\centering
	\caption{Model performance comparison on the Massachusetts Roads dataset.}
	\label{table1}
	\resizebox{0.95\columnwidth}{!}{
	\begin{tabular}{c|ccccc}
		\hline
		Methods           & Accuracy         & Precision        & Recall           & F1\_Score        & mIoU             \\ \hline
		DeepLabv3+\cite{deeplabv3+} & 0.9763          & 0.7808          & 0.6903          & 0.7328          & 0.7769          \\
		DenseUNet\cite{denseunet} & 0.9393 & 0.7825 & 0.7041 & 0.7407 & 0.7447 \\
		D-LinkNet\cite{dlinknet} & 0.9795          & \textbf{0.8765} & 0.6549          & 0.7497          & 0.7892          \\
		Ours              & \textbf{0.9804} & 0.8254          & \textbf{0.7399} & \textbf{0.7803} & \textbf{0.8098}  \\ \hline
	\end{tabular}
	}
\end{table}

\begin{figure*}[htbp]
	\centering
	\subfigure[]{
		\begin{minipage}[t]{0.15\linewidth}
			\centering
			 \includegraphics[width=\linewidth]{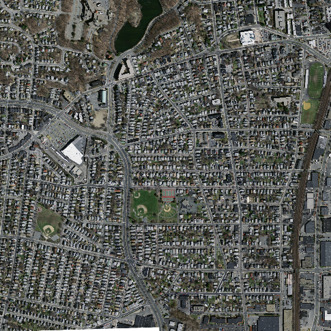}
			\\[5pt]
			 \includegraphics[width=\linewidth]{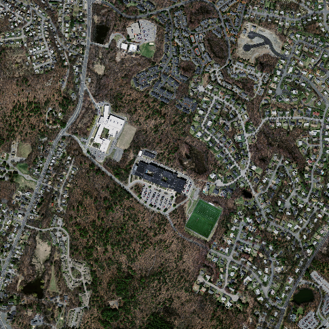}
			\\[5pt]
			 \includegraphics[width=\linewidth]{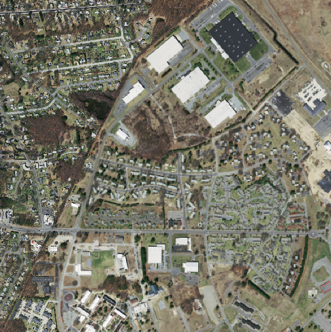}
			\vspace{-7pt}
		\end{minipage}	
	}
    \hspace{-0.01\linewidth}
	\subfigure[]{
		\begin{minipage}[t]{0.15\linewidth}
			\centering
			 \includegraphics[width=\linewidth]{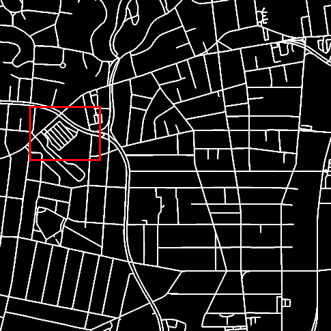}
			\\[5pt]
			 \includegraphics[width=\linewidth]{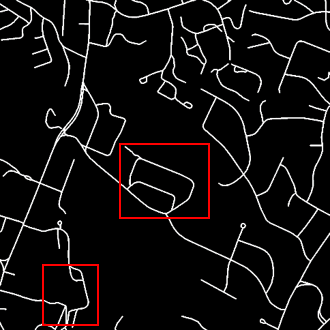}
			\\[5pt]
			 \includegraphics[width=\linewidth]{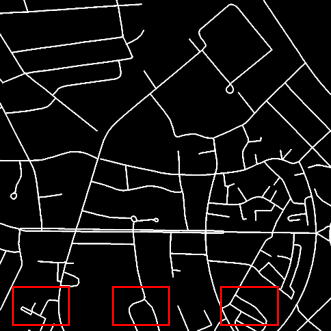}
			\vspace{-7pt}
		\end{minipage}
	}
    \hspace{-0.01\linewidth}
	\subfigure[]{
		\begin{minipage}[t]{0.15\linewidth}
			\centering
			 \includegraphics[width=\linewidth]{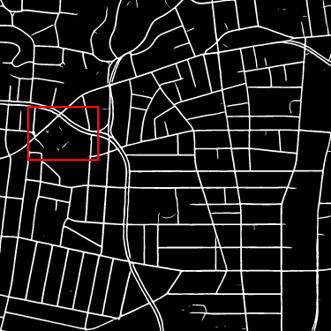}
			\\[5pt]
			 \includegraphics[width=\linewidth]{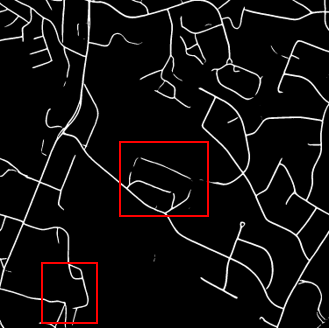}
			\\[5pt]
			 \includegraphics[width=\linewidth]{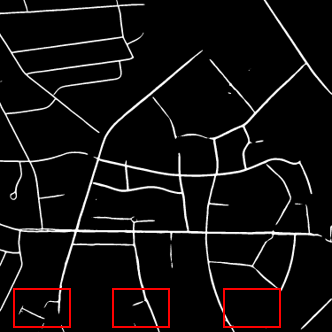}
			\vspace{-7pt}
		\end{minipage}
	}
    \hspace{-0.01\linewidth}
	\subfigure[]{
		\begin{minipage}[t]{0.15\linewidth}
			\centering
			 \includegraphics[width=\linewidth]{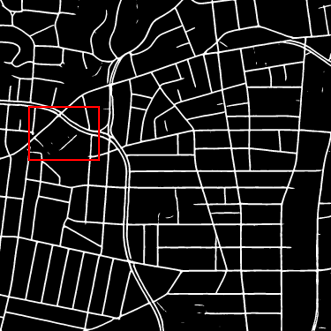}
			\\[5pt]
			 \includegraphics[width=\linewidth]{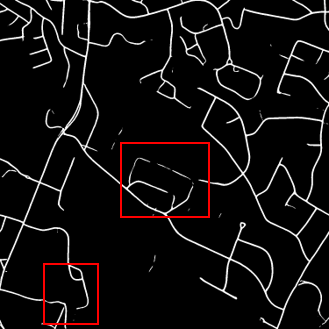}
			\\[5pt]
			 \includegraphics[width=\linewidth]{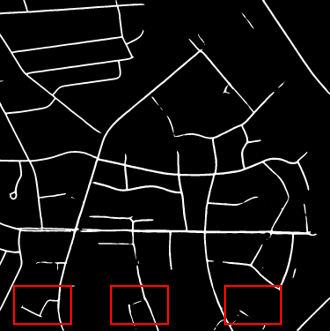}
			\vspace{-7pt}
		\end{minipage}
	}
    \hspace{-0.01\linewidth}
	\subfigure[]{
		\begin{minipage}[t]{0.15\linewidth}
			\centering
			 \includegraphics[width=\linewidth]{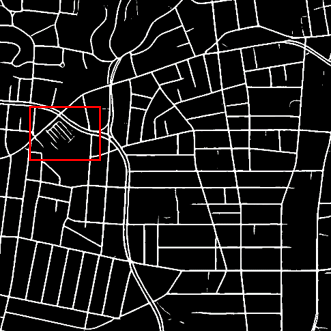}
			\\[5pt]
			 \includegraphics[width=\linewidth]{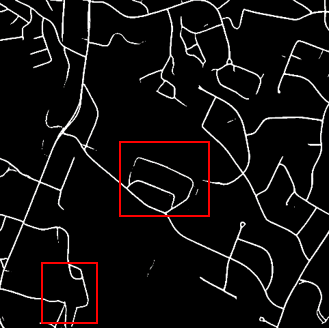}
			\\[5pt]
			 \includegraphics[width=\linewidth]{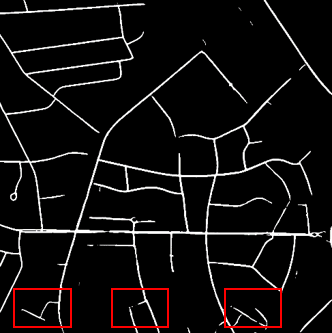}
			\vspace{-7pt}
		\end{minipage}
	}	
	\caption{Visual experimental results on the Massachusetts test set. (a) input image, (b) ground truth, (c) U-Net, (d) U-Net+DCAM, and (e) U-Net+DCAM+dual decoder.}
	\label{fig:figure5}
	\vspace{-1em}
\end{figure*}

Some visual experimental results are presented in Fig. \ref{fig:figure4}, where red frames highlight the easily missed parts due to the vegetation cover and the small size. From these results we can observe that our model has a better detection performance on small-sized roads than the reference ones. For example, roads in the parking lot are similar to the background in color (the first line), and some small roads covered by vegetation are thin and intermittent (the second and third lines). In these two cases, both reference models cannot effectively detect these small-sized roads, while our model offers a much higher detection rate. Similarly, compared to the reference models, our model shows much better continuity (the fourth line), a lower misjudgment rate, and a higher recall rate (the fifth and sixth lines) for small-sized roads.

Both the test metrics and visualization results show that DDU-Net can adequately capture the local and global context semantic information, and performs better than other comparison models for road extraction in complex environments, where multi-size roads coexist and the small-sized roads are prone to being missed due to the surrounding vegetation or building shadows. Since DDU-Net can retain more detailed information, small-sized roads are segmented more effectively with better continuity and smoother edges. Compared to mainstream road extraction models, DDU-Net achieves a lower misjudgment rate and a higher recall rate.

\begin{table}[t]
	\renewcommand\arraystretch{1.2}
	\centering
	\caption{Ablation results on the Massachusetts Roads dataset.}
	\label{table3}
	\resizebox{\columnwidth}{!}{
		\begin{tabular}{c|ccccc}	
			\hline
			Methods                                                   & Accuracy         & Precision        & Recall           & F1\_Score        & mIoU             \\ \hline
			U-Net                                                     & 0.9798          & \textbf{0.8333} & 0.7134          & 0.7687          & 0.8017          \\
			U-Net + DCAM                & 0.9800          & 0.8127          & \textbf{0.7467} & 0.7783          & 0.8082          \\
			U-Net + DCAM + Dual Decoder & \textbf{0.9804} & 0.8254          & 0.7399          & \textbf{0.7803} & \textbf{0.8098} \\ \hline
		\end{tabular}
	}
\end{table}

Further, we carry out ablation experiments to examine the effectiveness of the developed DCAM and the dual-decoder structure in the proposed DDU-Net model. An improved U-Net model is employed as the baseline, where the encoder replaces the four pooling layers by a pretrained ResNet-50 in the original U-Net model in order to improve the road segmentation performance. The ablation results are listed in Table \ref{table3}, from which we can observe that the introduction of the DCAM and dual-decoder structure is attributed to a better balance between precision and recall, and thus improve the F1 score and mIoU performances. Specifically, the DCAM can enhance the F1 score and mIoU by 1.0\% and 0.7\%, respectively, and the extra dual decoder structure can improve on the baseline model by 1.2\% and 0.8\%, respectively. As can be seen from the visual experimental results shown in Fig. \ref{fig:figure5}, the DCAM can extract the features of multi-scale receptive field and attention perception, so that it can better learn the context information, strengthen semantic representation and identify roads with vegetation shadows more effectively. Since the dual-decoder structure pays more attention to the restoration of detail features, it can refine the extraction results, improve connectivity, and better distinguish the roads of similar color to the background, such as the roads between parking lots. Obviously, the deliberately designed DCAM and dual-decoder structure are effective in enhancing road continuity of small-sized roads and improving road detection in complex environments.

\begin{figure}[htbp]
	\centering
	\subfigure[]{
		\begin{minipage}[t]{0.23\linewidth}
			\centering
			\includegraphics[width=\linewidth]{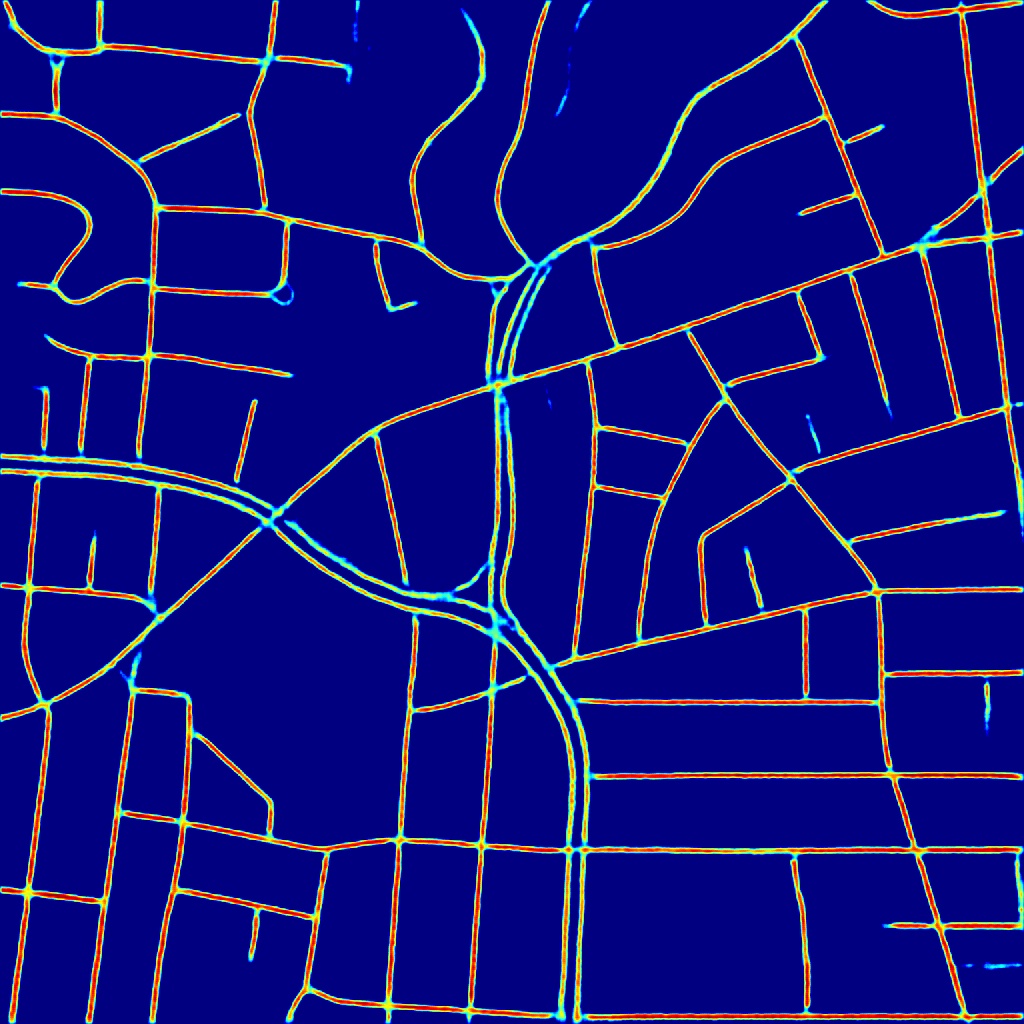}
			\\[5pt]
			 \includegraphics[width=\linewidth]{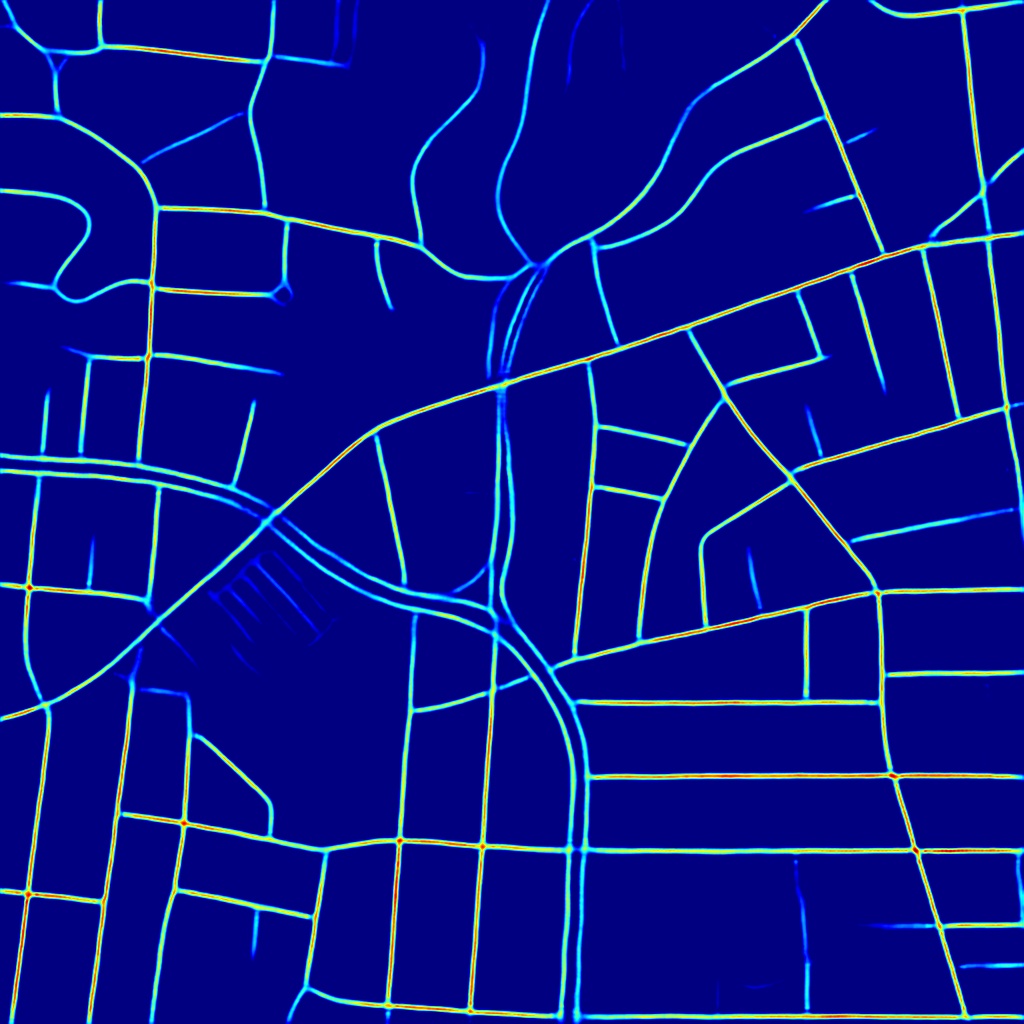}
			\vspace{-7pt}
		\end{minipage}	
	}
    \hspace{-0.035\linewidth}
	\subfigure[]{
		\begin{minipage}[t]{0.23\linewidth}
			\centering
			\includegraphics[width=\linewidth]{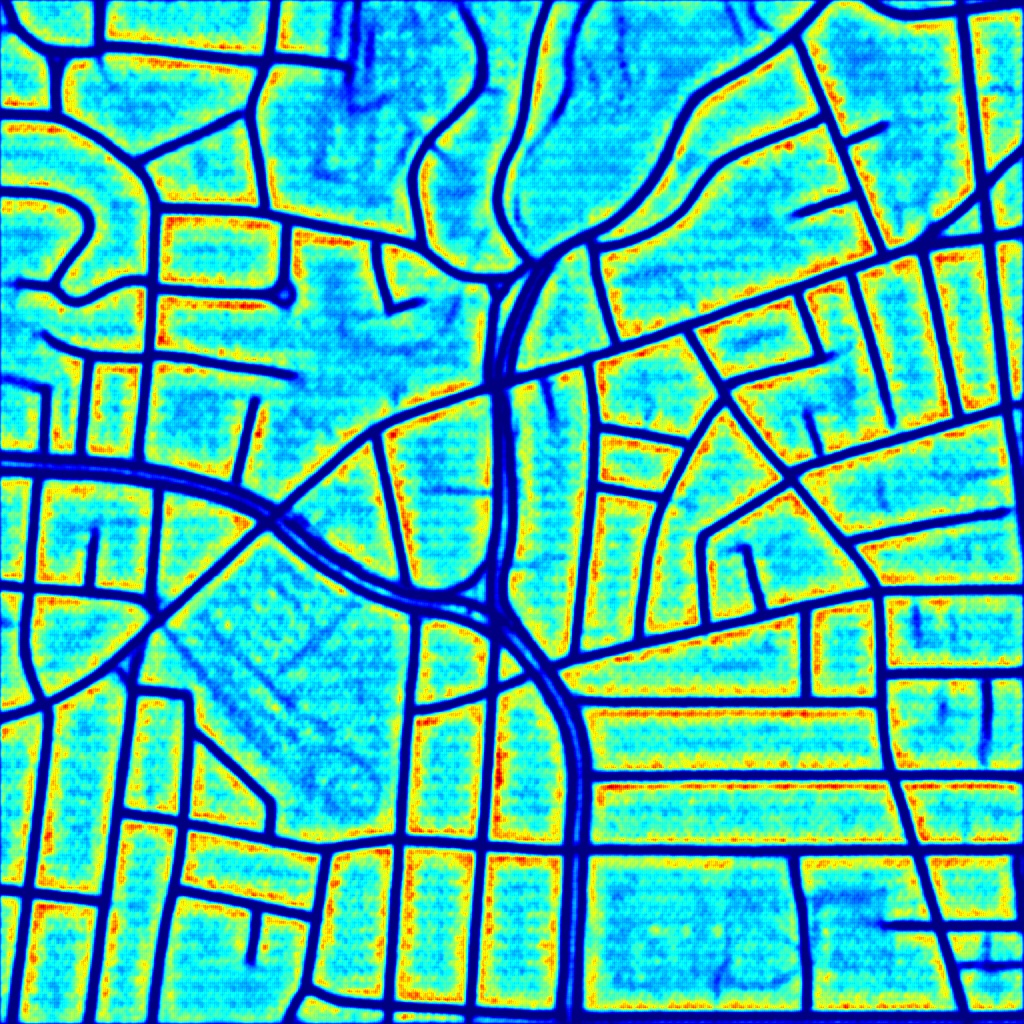}
			\\[5pt]
			 \includegraphics[width=\linewidth]{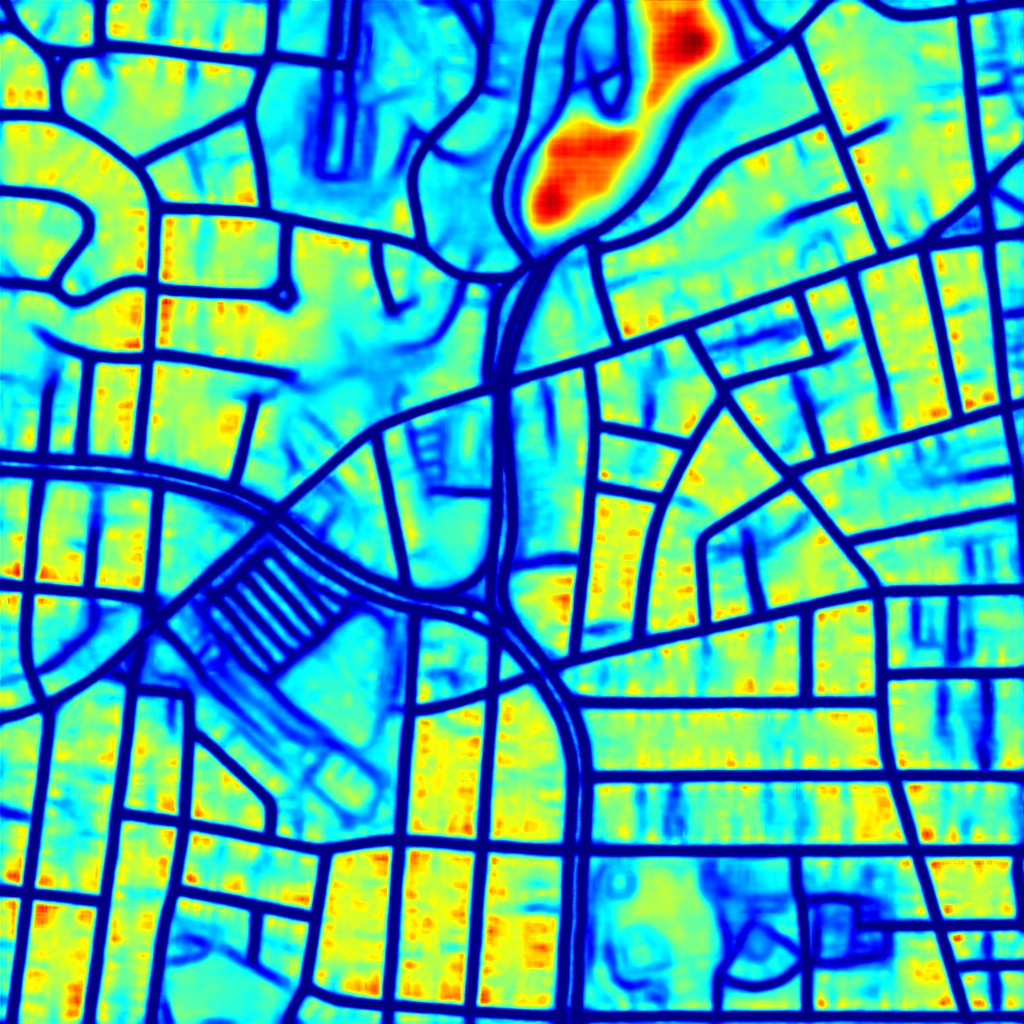}
			\vspace{-7pt}
		\end{minipage}
	}
    \hspace{-0.035\linewidth}
	\subfigure[]{
		\begin{minipage}[t]{0.23\linewidth}
			\centering
			\includegraphics[width=\linewidth]{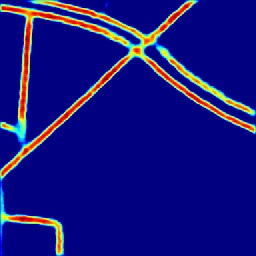}
			\\[5pt]
			 \includegraphics[width=\linewidth]{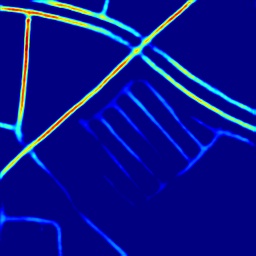}
			\vspace{-7pt}
		\end{minipage}
	}
    \hspace{-0.035\linewidth}
	\subfigure[]{
		\begin{minipage}[t]{0.23\linewidth}
			\centering
			\includegraphics[width=\linewidth]{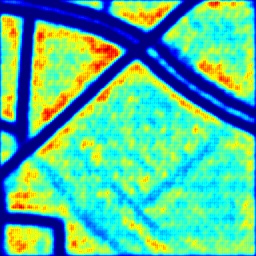}
			\\[5pt]
			 \includegraphics[width=\linewidth]{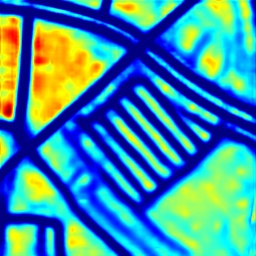}
			\vspace{-7pt}
		\end{minipage}
	}
	
	\caption{Feature visualization in the form of heatmap of the last layer of the feature map in D-LinkNet (1st lines) and our model (2nd lines). (a) feature of roads with label 1, (b) feature of background with label 0, (c) zoomed in road features, and (d) zoomed in background features. Each image shows the results of weighted average of all channels in the last layer of the feature map, and the brightness level represents the activation value.}
	\label{fig:figure6}
	\vspace{-1em}
\end{figure}
	
\begin{figure}[htbp]
	\centering
	\subfigure[]{
		\begin{minipage}[t]{0.23\linewidth}
			\centering
			 \includegraphics[width=\linewidth]{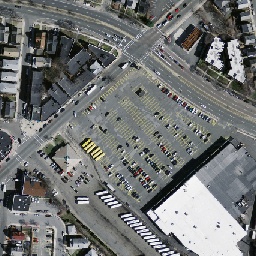}
			\vspace{-7pt}
		\end{minipage}	
	}
    \hspace{-0.035\linewidth}
	\subfigure[]{
		\begin{minipage}[t]{0.23\linewidth}
			\centering
			 \includegraphics[width=\linewidth]{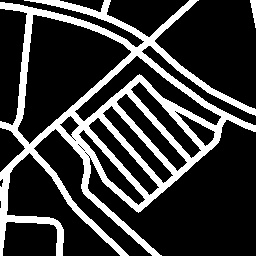}
			\vspace{-7pt}
		\end{minipage}
	}
    \hspace{-0.035\linewidth}
	\subfigure[]{
		\begin{minipage}[t]{0.23\linewidth}
			\centering
			 \includegraphics[width=\linewidth]{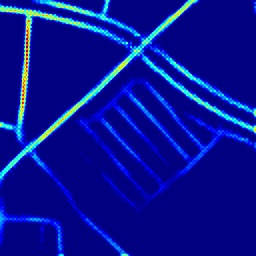}
			\vspace{-7pt}
		\end{minipage}
	}
    \hspace{-0.035\linewidth}
	\subfigure[]{
		\begin{minipage}[t]{0.23\linewidth}
			\centering
			 \includegraphics[width=\linewidth]{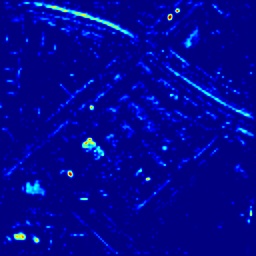}
			\vspace{-7pt}
		\end{minipage}
	}
	\caption{Feature visualization of the output feature maps of the two decoder in our model. (a) Original image, (b) ground truth, (c) road features of the large decoder, and (d) road features of the small decoder.}
	\label{fig:figure7}
	\vspace{-1em}
\end{figure}

Furthermore, Fig. \ref{fig:figure6} visualizes the feature map of the last layer in the form of heatmap for D-LinkNet and our model. We can observe from Fig. \ref{fig:figure6} the following: 
\begin{itemize}
\item The activation graphs in (b) and (d) show our model is capable of learning more abundant background features, containing more detailed information, and getting more comprehensive semantic representation, which is very helpful for road extraction in complex background environments;
\item The graphs in (a) and (c) show that our model can extract more road features with much better accuracy and continuity, especially for small-sized roads, which further validates the ability of our model in learning less redundant information and integrating global context information.
\end{itemize}

Moreover, in order to examine the effectiveness of the dual decoder structure in our model, we also visualize the output feature maps of the two decoders. As shown in Fig. \ref{fig:figure7} (c), the heatmap of the output feature map of the decoder has a lower brightness level representing small activation values corresponding to less continuous roads. But the general trend of the road can be seen from the large decoder's output feature, which shows that the large decoder attaches more importance to the integration of semantic information according to context dependency. The output features of the small decoder in (d) are scattered, and more attention is paid to the details of the roads like road edges, road markings in the lane, which help improve road continuity and segment the roads more accurately and smoothly. Thus, it can be concluded that the large decoder is capable of integrating local and global information, while the small decoder pays more attention to details and complements the final classification results.

\section{Conclusion}
In this paper, the DDU-Net model was proposed to greatly enhance road extraction from HRSIs. In the proposed DDU-Net model, the dual-decoder structure and DCAM were deliberately designed for small-sized road extraction where multi-sized roads coexist. In the proposed model, the introduced DCAM uses dilated convolution for receptive field expansion and multi-scale feature fusion, and employs a spatial channel attention mechanism to achieve attention-aware features. The designed dual-decoder structure can better capture global context information and retain more lower-level features. As a consequence, it can recover more details in the segmentation mask and improve the performance of road extraction in complex environments. Comprehensive performance evaluations were performed on the open Massachusetts Roads dataset, and comparisons were conducted with the state-of-the-art DeepLabv3+, DenseUNet, and D-LinkNet models. Our experimental results showed that the proposed DDU-Net outperforms all reference models, especially for small-sized roads covered by vegetation and complex background. Through ablation experiments, the effectiveness of the proposed DCAM and dual decoder structure were also validated. The features of the HRSIs were also visualized in the form of heatmap to demonstrate that the proposed model is capable of better learning both global and detailed features.

\ifCLASSOPTIONcaptionsoff
  \newpage
\fi



\bibliographystyle{IEEEtran}
\bibliography{reference}
\end{document}